\documentclass{article}

\usepackage[preprint]{neurips_2021}

\usepackage[utf8]{inputenc} 
\usepackage[T1]{fontenc}    
\usepackage{hyperref}       
\usepackage{url}            
\usepackage{booktabs}       
\usepackage{amsfonts}       
\usepackage{nicefrac}       
\usepackage{microtype}      
\usepackage{xcolor}         

\usepackage{bm}
\usepackage{multirow}
\usepackage{indentfirst}
\usepackage{times}
\usepackage{epsfig}
\usepackage{graphicx}
\usepackage{amsmath}
\usepackage{amssymb}
\usepackage{bm}
\usepackage{graphicx}
\usepackage{booktabs}
\usepackage{bm}
\usepackage{multirow}
\usepackage{makecell}
\usepackage{booktabs}
\usepackage{amssymb}
\usepackage{subfigure}   
\usepackage{wrapfig}
\usepackage{verbatim}
\usepackage[mathscr]{eucal}
\usepackage{wrapfig}
\usepackage{subfigure}

\title{Cross-view Geo-localization with Evolving Transformer}

\author{%
  Hongji~Yang, Xiufan~Lu, Yingying~Zhu\thanks{Corresponding author.}\\
  College of Computer Science and Software Engineering,\\
  Shenzhen University\\
  \texttt{\{2070276033, luxiufan2019\}@email.szu.edu.cn, zhuyy@szu.edu.cn} \\
}

\begin{document}

\maketitle

\begin{abstract}
  In this work, we address the problem of cross-view geo-localization, which estimates the geospatial location of a street view image by matching it with a database of geo-tagged aerial images. The cross-view matching task is extremely challenging due to drastic appearance and geometry differences across views. Unlike existing methods that predominantly fall back on CNN, here we devise a novel evolving geo-localization Transformer (EgoTR) that utilizes the properties of self-attention in Transformer to model global dependencies, thus significantly decreasing visual ambiguities in cross-view geo-localization. We also exploit the positional encoding of Transformer to help the EgoTR understand and correspond geometric configurations between ground and aerial images. Compared to state-of-the-art methods that impose strong assumption on geometry knowledge, the EgoTR flexibly learns the positional embeddings through the training objective and hence becomes more practical in many real-world scenarios. Although Transformer is well suited to our task, its vanilla self-attention mechanism independently interacts within image patches in each layer, which overlooks correlations between layers. Instead, this paper propose a simple yet effective self-cross attention mechanism to improve the quality of learned representations. The self-cross attention models global dependencies between adjacent layers, which relates between image patches while modeling how features evolve in the previous layer. As a result, the proposed self-cross attention leads to more stable training, improves the generalization ability and encourages representations to keep evolving as the network goes deeper. Extensive experiments demonstrate that our EgoTR performs favorably against state-of-the-art methods on standard, fine-grained and cross-dataset cross-view geo-localization tasks. Codes are available in the supplementary material.
\end{abstract}

\section{Introduction}
\label{sec:introduction}

Estimating the geospatial location of a given image is of paramount importance for robot navigation~\cite{robot}, 3D reconstruction~\cite{3dReconstruct} and autonomous driving~\cite{self-drive}. Recently, cross-view geo-localization, which aims to match query ground images with geo-tagged database aerial/satellite images, has emerged as a promising proposal to address this problem. 
Despite its appealing application prospect, the cross-view matching task is extremely challenging due to drastic viewpoint changes between ground and aerial images. Thus, it is critical to understand and correspond both image content (appearance and semantics) and spatial layout across views. 

Towards the above goal, several recent works incorporate convolutional neural networks (CNNs) with NetVlad layers~\cite{cvm}, capsule networks~\cite{geocaps} or attention mechanisms~\cite{reweighting,safa} to learn visually discriminative representations. However, the locality assumption of their CNN architectures hinders their performance in complex scenarios, where visual interferences such as obstacles and transient objects (\textit{e}.\textit{g}., cars and pedestrian) may exist. Instead, human visual system utilizes not only \emph{local} information but also \emph{global} context to make more accurate predictions when visual signals are ambiguous or incomplete. Another branch of works exploits geometry prior knowledge to reduce ambiguities caused by geometric misalignments. Though promising, these methods either rely heavily on predefined orientation prior~\cite{orien}, or make a restrictive assumption that ground and aerial images are orientation-aligned~\cite{safa}. As a result, such strong assumptions limit the applicability of these approaches, which prompts us to seek a more flexible approach for encoding position-aware representations. 

Motivated by these observations, we introduce Transformer~\cite{transformer}, which exceeds in global contextual reasoning and thus can be naturally employed to reduce visual ambiguities in cross-view geo-localization. Besides, the positional encoding of Transformer enables our network to flexibly learn position-dependent representations.
Specifically, our proposed evolving geo-localization Transformer (EgoTR) is built upon two independent Vision Transformer (ViT)~\cite{vit} branches, which split a feature map into several sub-patches while modeling interactions between arbitrary patches. We show in the experiment that due to its context- and position-dependent natures, such a Transformer-based network is a well suited candidate for cross-view geo-localization and shows its superiority compared to the dominant CNN-based counterparts.

We also take a deep look at self-attention map, which is the integral part of Transformer and is independently learned in each Transformer block. Nevertheless, such an independent learning strategy overlooks correlations between layers. Specifically, relating features from adjacent layers could improve the representation ability of network~\cite{cross-layer}. Compared to simply fusing features from all layers, interacting between features from adjacent layers produces less information redundancy and noise interference as they are in closer abstraction level. To explore cross-layer correlations, we replace the self-attention with a novel self-cross attention mechanism. Simple yet effective, the proposed self-cross attention learns pairwise similarities between features of adjacent blocks rather than of the same block. On one hand, such a cross-block interaction strategy eases the information flow across Transformer blocks, thus leading to more stable and effective network optimization. On the other hand, the self-cross attention map not only models intra-relations of an image, but also reflects how the learned representations evolve in the previous block. As shown in the experiments, this could facilitate representations to keep evolving in deep layers, thus improving the quality of image representation. 

The key contributions of this work are as follows.

\begin{itemize}
    \item To the best of our knowledge, the EgoTR is the first model using Transformer for cross-view geo-localization. The globally context-aware nature of the EgoTR effectively reduces visual ambiguities in cross-view geo-localization, while the positional encoding endows the EgoTR with the notion of geometry, thus decreasing ambiguities caused by geometry misalignments. Since the position embeddings are learned without imposing strong assumption on the position knowledge, the EgoTR has wider practical applicability compared with state-of-the-art models.
    
    \item We propose a novel self-cross attention mechanism, which interacts within cross-layer patches to ensure effective information flow across Transformer blocks and to encourage representations to keep evolving. This simple yet effective design consistently enhances the representation and the generalization ability of the EgoTR, without adding additional computational cost.
    
    \item Extensive experiments demonstrate that our EgoTR brings consistent and significant performance improvements for a wide range of cross-view matching tasks, including standard, fine-grained and cross-dataset cross-view geo-localization. On all these tasks, the EgoTR exhibits its superiority of learning visually discriminative and position-aware representations and achieves a new state-of-the-art performance.
\end{itemize}

\section{Related Work}
The key to cross-view geo-localization is to understand and correspond both image content (appearance and semantics) and spatial layout across views. To this end, existing cross-view geo-localization methods can be roughly grouped to two categories: content-based and geometry-based. 

\textbf{Content-based methods} focus on learning image representations that are discriminative enough to distinguish between similar looking images. Leveraging on the success of CNNs, Workman and Jacobs~\cite{workman} first introduce CNNs to the cross-view matching task. Later on, Hu \emph{et al.}~\cite{cvm} incorporate a two-branch VGG~\cite{vgg} backbone network with NetVlad layers~\cite{geo-localization-image2} to learn viewpoint-invariant representations. They also devise a weighted soft-margin triplet loss, which can speed up the network training. Sun \emph{et al.}~\cite{geocaps} apply the powerful ResNet~\cite{resnet} as backbone networks. Coupled with capsule layers~\cite{capsnet}, their proposed GeoCapsNet is capable of modeling high-level semantics. To steer where to focus in images, the attention mechanism is introduced to the field of cross-view geo-localization. Cai \emph{et al.}~\cite{reweighting} introduce a lightweight attention module that combines spatial and channel attention mechanisms to emphasize visually salient features. They also propose a novel reweighting loss that adaptively allocates weights to triplets according to their difficulties, thus improving the quality of network training. SAFA~\cite{safa} employs a multi-head spatial attention module to aggregate informative and diverse embedding maps. While promising, few of the above methods pay enough attention to the global dependencies of cross-view images, which hinders the discriminativeness of their embedded features. Different from existing methods, this work makes the first exploration to introduce Transformer~\cite{transformer} to cross-view geo-localization. We demonstrate the importance of considering global dependencies for reducing visual ambiguities.

\textbf{Geometry-based methods} aim to correspond geometric configurations between ground and aerial images, which helps to reduce ambiguities caused by geometry misalignments. To this end, Liu and Li~\cite{orien} explicitly inject per-pixel orientation information into the network. Nevertheless, this is based on the assumption of accessibility of the groundtruth orientation, which is not always satisfied in practice. Shi \emph{et al.}~\cite{safa} employ polar transform algorithm to warp satellite images so that aerial images are geometrically aligned with ground images. However, this method is only applicable to the ideal case where the ground images are orientation-aligned panoramas. Even though the dynamic similarity module proposed in \cite{where_looking} overcomes this limitation, the bruteforce warping strategy of the polar transform overlooks the depth of the scene content and results in obvious appearance distortions, which hinders the performance improvement. 
Regmi and Shah~\cite{regmi} attempt to tackle this problem by synthesizing the corresponding satellite image from a ground query using conditional GANs (cGANs), but the synthesized images are always granulated and lack details. In this paper, our EgoTR explicitly encodes learnable positional embeddings into the network, without imposing strong assumption. Different from the previous works~\cite{orien,safa,where_looking}, as shown in the experiments, our EgoTR not only learns relative positional information but also considers the scene context when corresponding geometric configurations across views.

\section{Method: EgoTR}

In this paper, we propose a novel evolving geo-localization Transformer (EgoTR) architecture with the self-cross attention mechanism for cross-view geo-localization. The following sections detail our problem setting, objective, the EgoTR architecture and our proposed self-cross attention.

\subsection{Problem Formulation and Objective}
The goal of cross-view geo-localization is to localize a query ground image by matching it with a set of geo-tagged aerial images. We formulate this problem in the same way as prior works~\cite{orien,optimal,safa,where_looking}. 

Assume we have a training set $D=\{(g_1,a_1),..., (g_N,a_N)\}$ containing $N$ cross-view image pairs of ground images $g$ and aerial images $a$. To simplify the problem, during training phase, let each ground image $g_i$ corresponds to only one ground-truth aerial image $a_i$ ($i\in\{1,2,...,N\}$). 
Given a cross-view image pair $(g_i,a_i)$, we infer the corresponding image representations as $({\bf{F}}^g_i,{\bf{F}}^a_i)$. Then, the weighted soft-margin triplet loss~\cite{cvm} $L$ of the $i^{th}$ exemplar can be defined as follows:
\begin{equation}
    L = log(1+e^{\alpha (d({\bf{F}}^g_i,{\bf{F}}^a_i)-d({\bf{F}}^g_i,{\bf{F}}^a_j))})
\end{equation}

where $j\in\{1,2,...,N\}$ and $j\neq i$. $\alpha$ is a hyperparameter used to speed up training convergence, and $d(\cdot,\cdot)$ denotes the $L_2$ distance.

\begin{figure*}
\centering
\includegraphics[width=0.96\textwidth]{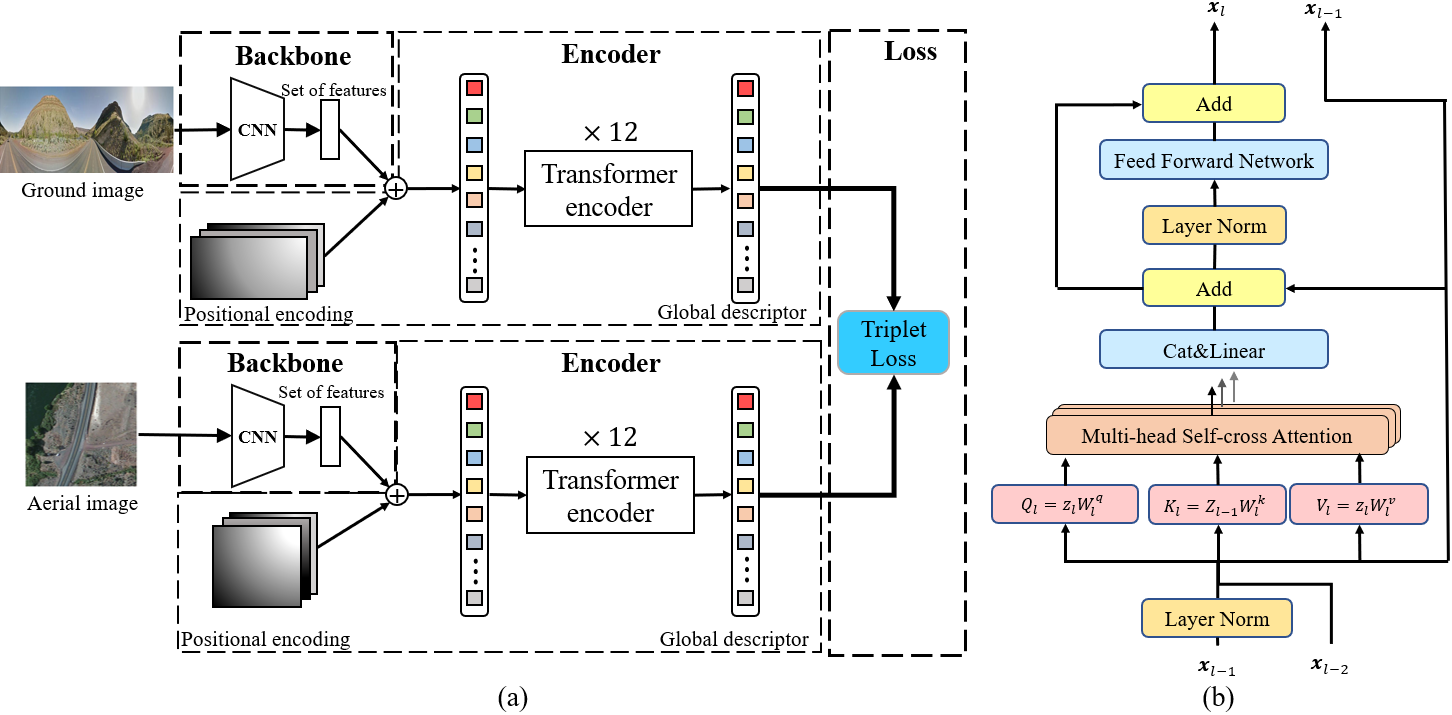}
\caption{(a) Overview of our evolving geo-localization Transformer (EgoTR). (b) Illustration of the encoder layer with the self-cross attention in the EgoTR. ${\bf x}_{l-1}$ denotes the input of layer $l$.} \vspace{-5pt}
\label{fig:network}
\end{figure*}

\subsection{Transformer for Cross-view Geo-localization}
We seek to develop an EgoTR architecture that explores the global context and the positional information of cross-view images.

\textbf{Preliminaries: Vision Transformer.} We first describe the Vision Transformer (ViT)~\cite{vit} architecture as background. Given an image, the ViT first splits it into several patches. Then, the ViT receives as input a sequence of linear projected patch embeddings ${\bf{x}}\in\mathbb{R}^{N\times D}$, where $N$ is the number of patches, and $D$ is the patch embedding size. After prepending a learnable class embedding ${\bf{x}}_{class}\in\mathbb{R}^{D}$, whose state at the output of the ViT is the image representation, and adding positional embeddings ${\bf{x}}_{pos}$ to ${\bf{x}}$, we gain ${\bf{x}}_0=[{\bf{x}}_{class};{\bf{x}}]+{\bf{x}}_{pos}$ and feed it into a $L$-layer Transformer encoder. 
Each layer consists of a Multihead Self-Attention module (MSA), Feed Forward Networks (FFN) and LayerNorm blocks (LN). 
Note that, the MSA is consist of multiple self-attention heads and a linear projection block. In order to make a clear comparison with our proposed self-cross attention head, we denote the input of layer $l$ ($l\in\{1,...,L\}$) as ${\bf x}_{l-1}$ and formulate a single self-attention head, the core of the vanilla MSA, as follows:
\begin{equation}
    \begin{split}
        {\bf z}_l&=LN({\bf{x}}_{l-1})\\
        {\bf Q}_l&={\bf z}_l{\bf W}^q_l, {\bf K}_l={\bf z}_l{\bf W}^k_l, {\bf V}_l={\bf z}_l{\bf W}^v_l\\
        {\bf A}_l&=softmax(\frac{{\bf Q}_l{\bf K}^T_l}{\sqrt{D}}){\bf V}_l
    \end{split}
    \label{self-att}
\end{equation}
where ${\bf W}^q_l$, ${\bf W}^k_l$ and ${\bf W}^v_l$ are linear projection matrices. 

\textbf{Domain-specific Transformer.}
The drastic domain gap between ground and aerial images hints that it is difficult to match the ground and aerial representations in the same data space. To suit the cross-view geo-localization task, we adopt a domain-specific Siamese-like architecture with two independent ViT branches of the same structure to separately learn ground and aerial image representations. The network overview is illustrated in Figure~\ref{fig:network} (a). Each branch is a hybrid structure consisting of a ResNet backbone extracting CNN feature map from an image input and a ViT modeling global context from the CNN feature map. The linear projection of patch embedding in the ViT is applied to the CNN feature map by regarding each $1 \times 1$ feature as a patch. 

\textbf{Learnable positional embedding.} Geometric cue can greatly simplify the cross-view geo-localization task~\cite{orien,safa}. Instead of imposing a pre-defined orientation knowledge on the network, this paper applies an efficient and flexible way to endow the network with the notion of geometry.
Specifically, we use learnable 1D positional embeddings in the ViT, \emph{i.e.} ${\bf{x}}_{pos}\in\mathbb{R}^{(N+1)\times D}$. By adding the positional embeddings to the linear patch embeddings, the transformed features become position-dependent. Furthermore, since we do not impose any assumption on the position knowledge but learn it through our learning objective, our EgoTR has wider practical applicability. As shown in the experiment, incorporating the learnable positional embeddings helps to capture relative positional information, which generalizes better to orientation-unknown images than absolute positional information. In addition, our EgoTR takes the scene content into account when corresponding cross-view geometry, which is complementary to the polar transform~\cite{safa,where_looking} and leads to a better localization performance.

\subsection{Self-cross Attention} 
In the vanilla ViT, the attention map is calculated independently in each layer. However, as mentioned before, such an independent learning strategy hinders the model's representation ability.
To improve the quality of the learned representations, this paper proposes a novel self-cross attention mechanism to interact features between adjacent layers. Specifically, the attention map of layer $l$ is learned not only based on ${\bf x}_{l-1}$, but also ${\bf x}_{l-2}$. Formally, in layer $l$, the self-cross attention can be represented as:
\begin{equation}
    \begin{split}
        {\bf z}_l&=LN({\bf{x}}_{l-1}), {\bf z}_{l-1}=LN({\bf{x}}_{l-2}),\\
        {\bf Q}_l&={\bf z}_l{\bf W}^q_l, {\bf K}_l={\bf z}_{l-1}{\bf W}^k_l, {\bf V}_l={\bf z}_l{\bf W}^v_l\\
        {\bf A}_l&=softmax(\frac{{\bf Q}_l{\bf K}^T_l}{\sqrt{D}}){\bf V}_l
    \end{split}
    \label{self-cross-att}
\end{equation}
Note that, for $l=1$, we set ${\bf z}_{l-1}=LN({\bf{x}}_{l-1})$. In Figure~\ref{fig:network} (b), we illustrate the structure of the self-cross attention-based encoder layer.

\begin{wrapfigure}{r}{6.5cm}
\vspace{-20pt}
\centering
\includegraphics[width=0.44\textwidth]{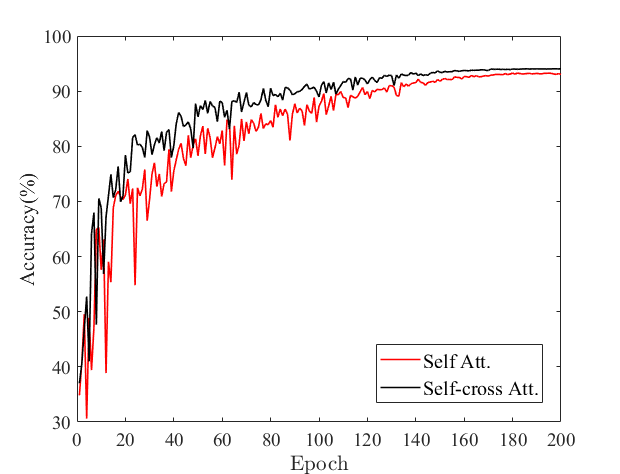}
\caption{Recall accuracy curve on CVUSA test set with increasing training epochs.}
\vspace{-10pt}
\label{fig:vs-epoch}
\end{wrapfigure}

\textbf{How the self-cross attention affects feature learning.}
Compared to the self-attention in Eq.~\ref{self-att}, our proposed self-cross attention creates short path between adjacent layers, thus allowing information flow effectively across layers. In fact, this shares the similar spirit with the ResNet~\cite{resnet}. To investigate how this affects our Transformer-based network, we plot the recall accuracy curve on CVUSA test set~\cite{cvusa} with increasing training epochs in Figure~\ref{fig:vs-epoch}. As shown, during the early training stage, the localization performance of the self-attention-based model fluctuates a lot, while our EgoTR exhibits more stable performance improvements as training processes. As a result, the stable training of the EgoTR could improves the network performance. Furthermore, by interacting cross-layer features, the attention map models how features evolve in the previous layer, which could decrease the representation similarity between layers, thus improving the network's representation ability. This is further discussed in the experiment.

\section{Experiment}
\label{sec:exp}

We first introduce three benchmark datasets we used to evaluate our EgoTR, evaluation protocols and implement details of our network. Then we compare our EgoTR with state-of-the-art models in Section~\ref{sec:compare-sota} and present ablation studies to illustrate the advantages of the proposed EgoTR in Section~\ref{sec:ablate}. Finally, we provide qualitative results in Section~\ref{sec:qualitative} to demonstrate the effectiveness of the positional embeddings in the EgoTR.

\begin{table}
	\caption{Comparisons with state-of-the-art models on CVACT\_val (standard cross-view geo-localization) and CVACT\_test (fine-grained geo-localization) datasets. Other results are quoted from \cite{safa} and \cite{where_looking}. ``PT'' indicates whether the model applies (w/) polar transform~\cite{safa} to aerial images or not (w/o).}
	\begin{center}
	\small
		\begin{tabular}{r|r|c|c|c|c|c|c|c|c|c} 
			\hline
			\multirow{3}{*}{PT}&\multirow{3}{*}{Model} & & \multicolumn{4}{c|}{CVACT\_val}& \multicolumn{4}{c}{CVACT\_test}\\ \cline{4-11}
			 ~& ~& Code& r@1 & r@5& r@10& r@1\%&  r@1 & r@5& r@10& r@1\%\\ 
			 ~ & ~ & Length& (\%) & (\%) & (\%) & (\%)& (\%) & (\%) & (\%) & (\%) \\\hline
			\multirow{5}{*}{w/o}& CVM-Net~\cite{cvm}& 4096 & 20.15 & 45.00 & 56.87 & 87.57&  5.41 & 14.79 & 25.63 & 54.53\\ 
			~ &Liu and Li~\cite{orien}& 1536& 46.96 & 68.28 & 75.48 & 92.01 & 19.21 & 35.97 & 43.30 & 60.69\\
			~ &CVFT~\cite{optimal}& 4096 & 61.05 & 81.33 & 86.52 & 95.93& 26.12 & 45.33 & 53.80 &71.69\\
			~ & SAFA~\cite{safa} & 4096 &78.28 & 91.60 & 93.79 & 98.15 & - & - & - & -\\
			~ &EgoTR &\textbf{768} &\textbf{83.14} &\textbf{93.84}&\textbf{95.51}&\textbf{98.40}& \textbf{58.33}& \textbf{84.23}& \textbf{88.60}& \textbf{95.83}\\ \hline
			\multirow{3}{*}{w/}&SAFA~\cite{safa}& 4096& 81.03 & 92.80 & 94.84 & 98.17& 55.50 & 79.94 & 85.08 & 94.49 \\
			~&Shi \emph{et al.}~\cite{where_looking}& 4096& 82.49 & 92.44 & 93.99 & 97.32& 35.63 & 60.07 & 69.10 & 84.75\\ 
			~ &Polar-EgoTR & \textbf{768} &\textbf{84.89} &\textbf{94.59}&\textbf{95.96}&\textbf{98.37}& \textbf{60.72} & \textbf{85.85} & \textbf{89.88} & \textbf{96.12}\\ \hline
		\end{tabular}
	\end{center}
	\label{compare_on_cvact}
\end{table}

\subsection{Dataset and Evaluation Protocol}
\textbf{Dataset.} To verify the effectiveness our model, we conduct extensive experiments on three widely used benchmarks: CVUSA~\cite{cvusa} and CVACT~\cite{orien} (including CVACT\_val and CVACT\_test). CVUSA dataset provides 35,532 image pairs for
training and 8,884 image pairs for testing. CVACT dataset contains 35,532 pairs for training and 8,884 pairs for validation (denoted as CVACT\_val). Additionally, to support fine-grained city-scale geo-localization, CVACT also provides 92,802 image pairs with accurate geo-tags for testing (denoted as CVACT\_test).

\begin{wraptable}{r}{8.8cm}
    \vspace{-5pt}
	\caption{Comparisons with state-of-the-art methods on CVUSA~\cite{cvusa} dataset. For all the compared methods, we cite the results from \cite{where_looking} and \cite{safa} if not specialized.}
	\begin{center}
	\small
		\begin{tabular}{r|r|c|c|c|c} 
			\hline
			\multirow{2}{*}{PT} & \multirow{2}{*}{Model} &  r@1 & r@5& r@10& r@1\%\\ 
			~ & ~ & (\%) & (\%) & (\%) & (\%) \\\hline
			\multirow{10}{*}{w/o}& Workman \emph{et al.}~\cite{workman} & - & - &- &34.30 \\ 
			~ &Vo and Hays~\cite{vo}  & - & - &- &63.70 \\ 
			~ &Zhai \emph{et al.}~\cite{cvusa} & - & - &- & 43.20 \\ 
			~ &CVM-Net~\cite{cvm}& 22.47 & 49.98 & 63.18 & 93.62 \\ 
			~ &Liu and Li~\cite{orien} & 40.79 & 66.82 & 76.36 & 96.12 \\
			~ &Zheng \emph{et al.}~\cite{university} & 43.91 & 66.38 & 74.58 & 91.78 \\
			~ &Regmi and Shah~\cite{regmi}   & 48.75 & - & 81.27 & 95.98 \\
			~ &Siam-FCANet~\cite{reweighting}& - & - & - & 98.30 \\ 
			~ &CVFT~\cite{optimal} & 61.43 & 84.69 & 90.49 & 99.02\\ 
			~ & SAFA~\cite{safa} & 81.15 & 94.23 & 96.85 & 99.49 \\
			~ &EgoTR  & \textbf{91.99} & \textbf{97.68} &\textbf{98.65} & \textbf{99.75} \\ \hline
			
			\multirow{3}{*}{w/}&SAFA~\cite{safa}  & 89.84 & 96.93 & 98.14 & 99.64\\
			~&Shi \emph{et al.}~\cite{where_looking}  & 91.93 & 97.50 & 98.54 & 99.67 \\ 
			~&Polar-EgoTR & \textbf{94.05} & \textbf{98.27} & \textbf{98.99} & \textbf{99.67} \\ \hline
		\end{tabular}
	\end{center}
	\label{tb:compare_sota_cvusa}
	\vspace{-30pt}
\end{wraptable}

\textbf{Evaluation protocol.} In line with \cite{cvm,orien,optimal,safa,where_looking}, we evaluate our model by recall accuracy at top $K$ (r@$K$ for short, $K\in\{1,5,10,1\%\}$), which represents the probability of correct match(es) ranking within the first $K$ results. Note that r@1\% means the recall accuracy at top 1\% of test set. For CVUSA and CVACT\_val, a query ground image corresponds to a single aerial image, while for CVACT\_test, aerial images that locate within 5 meters of the query ground image can be seen as the correct matches.

\subsection{Implementation Detail}
If not specified, the ground and aerial image size are set to $128\times 512$ and $256 \times 256$, respectively. We empirically set model depth $L$ to 12 and initialize our EgoTR with pretrained parameters on ImageNet~\cite{imagenet}. The model is trained using AdamW~\cite{adamW} with cosine learning rate schedule on a 32GB NVIDIA V100 GPU. The learning rate is set to 1e-4, the weight decay is chosen to 0.03 and the batch size is 32. For the weighted soft-margin triplet loss~\cite{cvm}, $\alpha$ is set to 10.

\subsection{Comparing EgoTR with State-of-the-art Models}
\label{sec:compare-sota}

Here we compare our method with several state-of-the-art methods on CVUSA~\cite{cvusa}, CVACT\_val~\cite{orien} and CVACT\_test~\cite{orien} datasets. 
Unlike state-of-the-art methods that predominantly fall back on CNN, our proposed EgoTR makes the first effort to introduce Transformer to the field of cross-view geo-localization to learn globally context- and position-aware representations. Below, we verify that our EgoTR exceeds in learning visually discriminative and position-aware representations, thus achieving outstanding performance in various cross-view geo-localization tasks. Note that for fair comparison with works \cite{safa, where_looking} that use polar transform~\cite{safa}, a kind of data pre-processing algorithm, we apply the same warping strategy to aerial images before feeding them into the network (denoted as Polar-EgoTR) when comparing with these works. In this case, ground and warped aerial images are resized to $128\times 512$.

\textbf{Standard cross-view geo-localization.}
We first evaluate our EgoTR on standard cross-view geo-localization. Table \ref{compare_on_cvact} and \ref{tb:compare_sota_cvusa} show experimental results on CVACT\_val and CVUSA datasets, respectively. From the results, we could conclude that our EgoTR significantly surpasses the competing approaches in learning visually discriminative representations and corresponding geometric configurations across views. In particular, without applying the polar transform, our EgoTR achieves r@1 of 83.14\% on CVACT\_val dataset compared to 78.28\% obtained by the second best method, while on CVUSA dataset the EgoTR surpasses the second best method by a significant margin of 10.84 points at r@1. Moreover, when applying the polar transform, which geometrically aligns cross-view images, our EgoTR outperforms the competing methods, gaining 84.89\% and 94.05\% on CVACT\_val and CVUSA, respectively. The results indicates that the EgoTR is capable of capturing visually discriminative features by modeling global context. 
Furthermore, we could also find that removing the polar transform algorithm leads to significant performance degradation in SAFA (\emph{${-4.21\%}$} on CVACT\_val and \emph{${-8.69\%}$} on CVUSA) while less of a degradation is noted in our EgoTR (\emph{${-1.75\%}$} on CVACT\_val and \emph{${-2.06\%}$} on CVUSA). Namely, the EgoTR does not have to rely excessively on the polar transform to establish cross-view geometric correspondence, which could save considerable image pre-processing time on large-scale datasets.
This is because, adding the positional embeddings to the linear patch embeddings enables the EgoTR to learn position-aware representations and to correspond geometric configurations between ground and aerial images. Additional qualitative evidence for this is provided in Section \ref{sec:qualitative}.

\begin{wraptable}{r}{9.5cm}
 \vspace{-5pt}
	\caption{Cross-dataset cross-view geo-localization. The results are gained by retraining and evaluating the compared models using the released codes provided by their authors.}
	\begin{center}
	\small
		\begin{tabular}{r|c|c|c|c|c} 
			\hline
			 \multirow{2}{*}{Model}& \multirow{2}{*}{Task}&r@1 & r@5& r@10& r@1\% \\ 
			 ~ & ~ & (\%) & (\%) & (\%) & (\%) \\\hline
			SAFA~\cite{safa}& \multirow{3}{*}{CVUSA$\rightarrow$CVACT}& 30.40 & 52.93 & 62.29 & 85.82\\ 
			Shi \emph{et al.}~\cite{where_looking}&~& 33.66 & 52.17 &59.74& 79.67 \\
			Polar-EgoTR &~ & \textbf{47.55} &\textbf{70.58}&\textbf{77.39}&\textbf{91.39}\\ \hline
			SAFA~\cite{safa}& \multirow{3}{*}{CVACT$\rightarrow$CVUSA}& 21.45 & 36.55 & 43.79 & 69.83\\ 
			Shi \emph{et al.}~\cite{where_looking}&~& 18.47 & 34.46 & 42.28 & 69.01 \\
			Polar-EgoTR &~ & \textbf{33.00} &\textbf{51.87}&\textbf{60.63}&\textbf{84.79}\\ \hline
		\end{tabular}
	\end{center}
	\label{tb:cross-dataset}
\end{wraptable}

\textbf{Fine-grained cross-view geo-localization.}
To evaluate the representation ability of our model, we verify the EgoTR on fine-grained cross-view geo-localization task. Specifically, we compare the EgoTR with state-of-the-art methods on the challenging large-scale CVACT\_test dataset, which is fully gps-tagged for accurate localization. Table \ref{compare_on_cvact} shows the experimental results. Our EgoTR performs consistently better than all the competitors, achieving 58.33\% and 60.76\% at r@1 without and with the polar transform, respectively. These results further demonstrate that our EgoTR has strong representation capability.

\begin{wraptable}{r}{8.3cm}
\vspace{-20pt}
	\caption{Ablation studies of the EgoTR.}
	\begin{center}
		\begin{tabular}{r|c|c|c|c} 
			\hline
		    \multirow{2}{*}{Model} &  r@1 & r@5& r@10& r@1\%\\ 
			~ & (\%) & (\%) & (\%) & (\%) \\\hline
			Polar-EgoTR& 94.05 & 98.27 & 98.99 & 99.67\\ 
			w/o self-cross att. & 93.26 & 97.91 & 98.78 & 99.68 \\ 
			w/o positional emb. & 90.90 & 97.48 & 98.40 & 99.62 \\ \hline
			EgoTR & 91.99 & 97.68 & 98.65 & 99.75 \\
			w/o positional emb. & 89.04 & 96.88 & 98.44 & 99.61 \\ \hline
		\end{tabular}
	\end{center}
	\label{tb:ablate}
\end{wraptable}

\noindent \textbf{Cross-dataset cross-view geo-localization.} In the context of cross-view geo-localization, the transferring performance determines whether a model could be practically usable for real-life scenarios, where a query image may
be dramatically different from the training ground images.
As CVACT and CVUSA datasets are collected from two different countries, they have distinctly different scene styles. Based on this observation, to verify the transferring performance of our model, we train the EgoTR on CVUSA dataset and
test it on CVACT\_val (denoted as CVUSA$\rightarrow$CVACT),
and vice versa. Results are reported in Table~\ref{tb:cross-dataset}. We could find that our EgoTR outperforms the second best model at r@1 by a large margin of 13.89 points on the CVUSA$\rightarrow$CVACT task, while achieves 33.00\% at r@1 compared to 21.45\% gained by the second best model on the CVACT$\rightarrow$CVUSA task. The transferring results demonstrate the outstanding generalization ability and practical applicability of our EgoTR.

\noindent \textbf{Comparison in terms of code length.} To further illustrate the advantage of our method, we compare the EgoTR with state-of-the-art methods in terms of image descriptor dimension (also called code length) in Table~\ref{compare_on_cvact}. We observe that our EgoTR has extremely short code length of 768, which is five times shorter than that of the SAFA~\cite{safa}, CVFT~\cite{optimal} and CVM-Net~\cite{cvm}. A shorter code length not only implies the effective information encoding capability of the EgoTR, but also means that the EgoTR provides an alternative with less storage space, lower computational complexity and shorter running time to cross-view geo-localization.

\subsection{Ablation Study}
\label{sec:ablate}

To investigate the effectiveness of the positional embeddings and the self-cross attention mechanism, we conduct ablation studies by considering three scenarios: 1) where the positional embeddings are removed, 2) where the polar transform is removed and 3) where the self-cross attention is replaced by the self-attention.

\begin{wraptable}{r}{10cm}
\vspace{-20pt}
	\caption{Few-shot cross-view geo-localization on CVUSA~\cite{cvusa}. We describe the number of training pairs of each subset and its proportion (Prop.) to the original CVUSA dataset.}
	\small
	\begin{center}
		\begin{tabular}{r|r|c|c|c|c|c}
			\hline
			Training & \multirow{2}{*}{Prop.} & \multirow{2}{*}{Model} &  r@1 & r@5& r@10& r@1\%\\ 
			Pairs & ~ & ~ & (\%) & (\%) & (\%) & (\%) \\\hline
			\multirow{2}{*}{7,106}& \multirow{2}{*}{20\%} &Polar-EgoTR & \textbf{76.01} & \textbf{90.67} & \textbf{94.01} & \textbf{98.85}\\ 
			~ & ~ & w/o self-cross att.   & 75.37 & 90.42 & 92.85 & 98.66\\\hline
			
			\multirow{2}{*}{14,212}&\multirow{2}{*}{40\%}& Polar-EgoTR  & \textbf{86.06} & \textbf{95.80} &  \textbf{97.16}&  99.38\\ 
			~ & ~ &w/o self-cross att.& 85.54 & 95.14 & 97.07 & \textbf{99.42} \\ \hline
			
			\multirow{2}{*}{21,319}& \multirow{2}{*}{60\%} & Polar-EgoTR  &  \textbf{90.30} & \textbf{96.96}  & \textbf{98.26}  &  \textbf{99.67}\\
			~ & ~& w/o self-cross att. & 88.74 & 96.60 & 98.09 & 99.66\\  \hline
			
			\multirow{2}{*}{35,532}& \multirow{2}{*}{100\%} & Polar-EgoTR & \textbf{94.05} & \textbf{98.27} & \textbf{98.99} & 99.67 \\ 
			~ & ~ &w/o self-cross att.& 93.26 & 97.91 & 98.78 & \textbf{99.68}\\ \hline
		\end{tabular}
	\end{center}
	\label{tb:ablate-few-shot}
\end{wraptable}

\textbf{Positional encoding.} We first analyze the importance of the positional embeddings and report the ablation studies on CVUSA dataset in Table~\ref{tb:ablate}. From the results, we can make the following observations. First, the positional encoding endows the network with the concept of position and yields consistent improvements. In particular, adding positional embeddings improves the r@1 performance of the EgoTR from 89.04\% to 91.99\% and brings 3.15 points improvement at r@1 to the Polar-EgoTR. Second, we could also find that the positional embeddings are complementary to the polar transform~\cite{safa,where_looking}. Specifically, combining the polar transform with our EgoTR improves the r@1 accuracy from 91.99\% to 94.05\% ($+2.06\%$), while adding the positional embeddings to Polar-EgoTR boosts the r@1 performance from 90.90\% to 94.05\% ($+3.15\%$). An interpretation of this is that, incorporating the learnable positional embeddings is capable of considering the scene content when corresponding cross-view geometric configurations, which is overlooked in polar transform algorithm as mentioned before. As a result, the polar transform and the positional encoding could jointly improve the network performance. In Section \ref{sec:qualitative}, we show qualitative results to verify this inference.

\begin{wrapfigure}{r}{6.5cm}
\vspace{-20pt}
\centering
\includegraphics[width=0.5\textwidth]{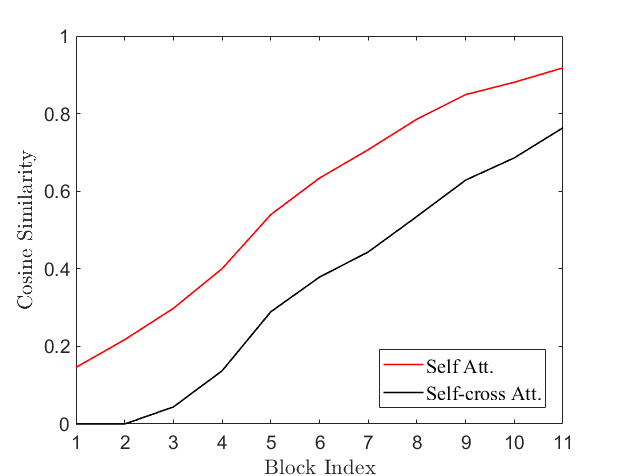}
\caption{Cross-layer similarity between the last layer and previous layers.}
\label{fig:vs-similarity}
\vspace{-10pt}
\end{wrapfigure}

\textbf{Self-cross attention.} In Table \ref{tb:ablate}, we ablate the self-cross attention mechanism by replacing it with the self-attention on CVUSA dataset. We can observe from Table \ref{tb:ablate} that, without imposing increase in model complexity, the self-cross attention mechanism improves r@1 performance from 93.26\% to 94.05\%, which manifests the effectiveness of the self-cross attention. Moreover, in Table~\ref{tb:ablate-few-shot}, we ablates the self-cross attention on few-shot cross-view geo-localization task. The few-shot task aims to learn a model that can achieve generalization from only a small number of training examples~\cite{few-shot}. To support this task, we randomly select a certain percentage (20\%/40\%/60\%) of samples from CVUSA dataset to generate three subsets. The size of each subset and its corresponding proportion to CVUSA dataset are illustrated in Table \ref{tb:ablate-few-shot}. Results show that, replacing the self-cross attention with the self-attention consistently harms the network performance on few-shot task. This indicates that, the self-cross attention not only improves network performance, but also enhances its generalization ability. Furthermore, we also investigate how the self-cross attention affects feature learning. 
In Figure~\ref{fig:vs-similarity}, we compare the final representation with the output of each intermediate layer by measuring their cosine similarity. As can be observed, replacing the self-attention with our proposed self-cross attention significantly and consistently decreases the representation similarity between layers. This result implies that the self-cross attention could prevent the learned representations of Transformer layers from being overly similar with each others and encourage the model to evolve its representations layer by layer, thus improving the performance of the final image representations. 

{\begin{figure}
\centering
    \subfigure[The ground PE of the EgoTR]{
    \label{fig:pos-grd}
    \includegraphics[width=0.23\textwidth]{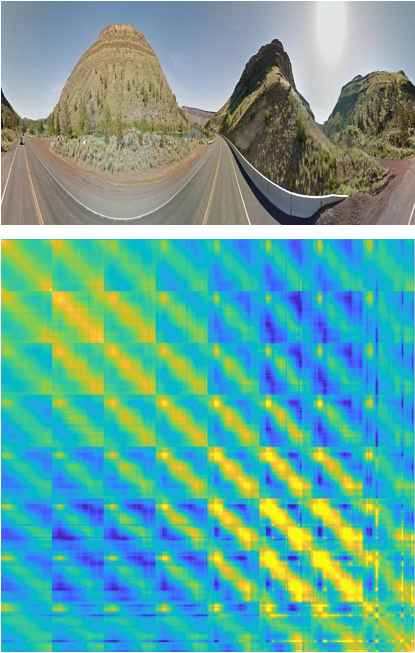}
    }
    \subfigure[The aerial PE of the EgoTR]{
    \label{fig:pos-sat}
    \includegraphics[width=0.23\textwidth]{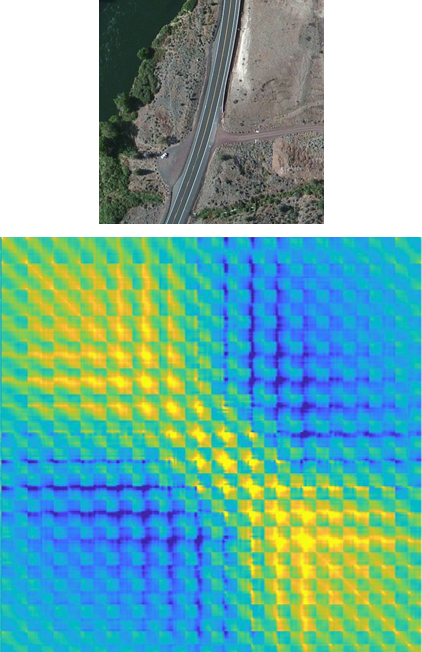}
    }
    \subfigure[The ground PE of the Polar-EgoTR]{
    \label{fig:pos-grd-polar}
    \includegraphics[width=0.23\textwidth]{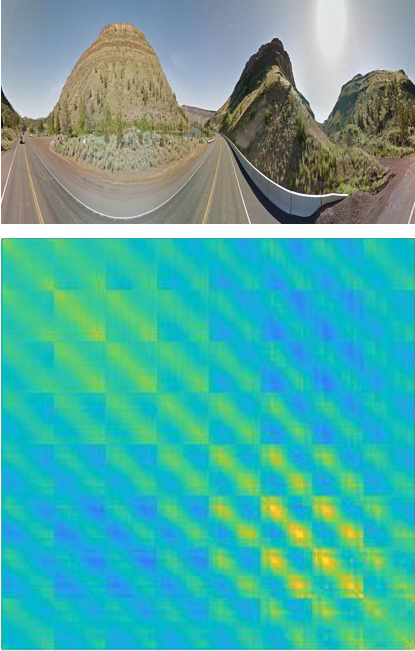}
    }
    \subfigure[The aerial PE of the Polar-EgoTR]{
    \label{fig:pos-sat-polar}
    \includegraphics[width=0.23\textwidth]{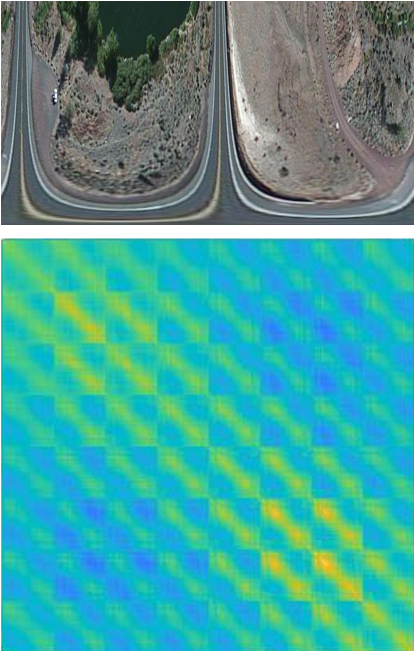}
    }
\caption{Dot products between the learnable positional embeddings (PE). Yellow indicates the two positional embeddings are closer. Better viewed in color and with zoom-in.} 
\label{fig:pos-embed}
\end{figure}}

\subsection{Qualitative Analysis}
\label{sec:qualitative}
We conduct detailed qualitative analysises on the learnable positional embeddings to investigate whether they encode and correspond geometric configurations across views and which positional information they can learn. To this end, we calculate dot products between two arbitrary positional embeddings of the EgoTR. From Figure \ref{fig:pos-grd} and \ref{fig:pos-sat}, we could find that each positional embedding is close to its neighbors with small location offsets. This implies that, incorporating the learnable positional embeddings captures relative positional information. 
Additionally, we could clearly observe that, the visualization maps of the EgoTR are distinctly different across views in Figure \ref{fig:pos-grd} and \ref{fig:pos-sat}, while the visualization maps of the Polar-EgoTR look similar to each other in Figure \ref{fig:pos-grd-polar} and \ref{fig:pos-sat-polar}. Such similar results of the Polar-EgoTR are reasonable, since cross-view images are geometrically aligned by the polar transform. This result further confirms that the positional encoding could capture cross-view geometric configurations.
Furthermore, in Figure~\ref{fig:pos-sat}, it is obvious that the positional vectors near the center of each aerial image are close to their neighbors that have fairly small distance offsets. In contrast, the positional embeddings far away from the center are close to their neighbors with relatively larger offsets. Such an offset difference is learned because of the geometric perspective of ground images. Specifically, two objects look farther away from each other in a ground image when they lie near the center of the corresponding aerial image, but look nearer to each other in the ground image when they lie far away from the center. 
This result unveils that the EgoTR could consider the scene content when corresponding geometric configurations across views.

\section{Conclusion and Future Work}
\label{sec:conclude}
In this paper, we propose a novel EgoTR architecture capable of learning globally context- and position-aware representations. We also propose a novel self-cross attention to facilitate information flow across layers and encourage representations to keep evolving as the network layer goes deeper. Extensive experiments demonstrate that the EgoTR outperforms state-of-the-art methods in standard, fine-grained and cross-dataset cross-view geo-localization tasks. In addition, we also conduct ablation studies and qualitative analyses to verify the effectiveness of the learnable positional embeddings and the self-cross attention. One main limitation of the EgoTR is its large demand on GPU memory. Moreover, the EgoTR is built on top of the pretrained Transformer, which requires a large amount of data for training. For future work, we aim to develop a data-efficient Transformer-based model with less memory consumption for cross-view geo-localization.

\bibliographystyle{abbrv}
\bibliography{reference.bib}

\begin{thebibliography}{25}
\providecommand{\natexlab}[1]{#1}
\providecommand{\url}[1]{\texttt{#1}}
\expandafter\ifx\csname urlstyle\endcsname\relax
  \providecommand{\doi}[1]{doi: #1}\else
  \providecommand{\doi}{doi: \begingroup \urlstyle{rm}\Url}\fi

\bibitem[Arandjelovic et~al.(2016)Arandjelovic, Gronat, Torii, Pajdla, and
  Sivic]{geo-localization-image2}
Relja Arandjelovic, Petr Gronat, Akihiko Torii, Tomas Pajdla, and Josef Sivic.
\newblock Netvlad: Cnn architecture for weakly supervised place recognition.
\newblock In \emph{Proceedings of the IEEE conference on computer vision and
  pattern recognition}, pages 5297--5307, 2016.

\bibitem[Cai et~al.(2019)Cai, Guo, Khan, Hu, and Wen]{reweighting}
Sudong Cai, Yulan Guo, Salman Khan, Jiwei Hu, and Gongjian Wen.
\newblock Ground-to-aerial image geo-localization with a hard exemplar
  reweighting triplet loss.
\newblock In \emph{Proceedings of the IEEE/CVF International Conference on
  Computer Vision}, pages 8391--8400, 2019.

\bibitem[Deng et~al.(2009)Deng, Dong, Socher, Li, Li, and Fei-Fei]{imagenet}
Jia Deng, Wei Dong, Richard Socher, Li-Jia Li, Kai Li, and Li~Fei-Fei.
\newblock Imagenet: A large-scale hierarchical image database.
\newblock In \emph{2009 IEEE conference on computer vision and pattern
  recognition}, pages 248--255. Ieee, 2009.

\bibitem[Dosovitskiy et~al.(2020)Dosovitskiy, Beyer, Kolesnikov, Weissenborn,
  Zhai, Unterthiner, Dehghani, Minderer, Heigold, Gelly, et~al.]{vit}
Alexey Dosovitskiy, Lucas Beyer, Alexander Kolesnikov, Dirk Weissenborn,
  Xiaohua Zhai, Thomas Unterthiner, Mostafa Dehghani, Matthias Minderer, Georg
  Heigold, Sylvain Gelly, et~al.
\newblock An image is worth 16x16 words: Transformers for image recognition at
  scale.
\newblock \emph{arXiv preprint arXiv:2010.11929}, 2020.

\bibitem[H{\"a}ne et~al.(2017)H{\"a}ne, Heng, Lee, Fraundorfer, Furgale,
  Sattler, and Pollefeys]{self-drive}
Christian H{\"a}ne, Lionel Heng, Gim~Hee Lee, Friedrich Fraundorfer, Paul
  Furgale, Torsten Sattler, and Marc Pollefeys.
\newblock 3d visual perception for self-driving cars using a multi-camera
  system: Calibration, mapping, localization, and obstacle detection.
\newblock \emph{Image and Vision Computing}, 68:\penalty0 14--27, 2017.

\bibitem[He et~al.(2020)He, Hong, Liu, Xu, Zha, and Wang]{few-shot}
Jun He, Richang Hong, Xueliang Liu, Mingliang Xu, Zheng-Jun Zha, and Meng Wang.
\newblock Memory-augmented relation network for few-shot learning.
\newblock In \emph{Proceedings of the 28th ACM International Conference on
  Multimedia}, pages 1236--1244, 2020.

\bibitem[He et~al.(2016)He, Zhang, Ren, and Sun]{resnet}
Kaiming He, Xiangyu Zhang, Shaoqing Ren, and Jian Sun.
\newblock Deep residual learning for image recognition.
\newblock In \emph{Proceedings of the IEEE conference on computer vision and
  pattern recognition}, pages 770--778, 2016.

\bibitem[Hu et~al.(2018)Hu, Feng, Nguyen, and Lee]{cvm}
Sixing Hu, Mengdan Feng, Rang~MH Nguyen, and Gim~Hee Lee.
\newblock Cvm-net: Cross-view matching network for image-based ground-to-aerial
  geo-localization.
\newblock In \emph{Proceedings of the IEEE Conference on Computer Vision and
  Pattern Recognition}, pages 7258--7267, 2018.

\bibitem[Liu and Li(2019)]{orien}
Liu Liu and Hongdong Li.
\newblock Lending orientation to neural networks for cross-view
  geo-localization.
\newblock In \emph{Proceedings of the IEEE/CVF Conference on Computer Vision
  and Pattern Recognition}, pages 5624--5633, 2019.

\bibitem[Loshchilov and Hutter(2018)]{adamW}
Ilya Loshchilov and Frank Hutter.
\newblock Fixing weight decay regularization in adam.
\newblock 2018.

\bibitem[McManus et~al.(2014)McManus, Churchill, Maddern, Stewart, and
  Newman]{robot}
Colin McManus, Winston Churchill, Will Maddern, Alexander~D Stewart, and Paul
  Newman.
\newblock Shady dealings: Robust, long-term visual localisation using
  illumination invariance.
\newblock In \emph{2014 IEEE international conference on robotics and
  automation (ICRA)}, pages 901--906. IEEE, 2014.

\bibitem[Middelberg et~al.(2014)Middelberg, Sattler, Untzelmann, and
  Kobbelt]{3dReconstruct}
Sven Middelberg, Torsten Sattler, Ole Untzelmann, and Leif Kobbelt.
\newblock Scalable 6-dof localization on mobile devices.
\newblock In \emph{European conference on computer vision}, pages 268--283.
  Springer, 2014.

\bibitem[Pang et~al.(2020)Pang, Zhao, Zhang, and Lu]{cross-layer}
Youwei Pang, Xiaoqi Zhao, Lihe Zhang, and Huchuan Lu.
\newblock Multi-scale interactive network for salient object detection.
\newblock In \emph{Proceedings of the IEEE/CVF Conference on Computer Vision
  and Pattern Recognition}, pages 9413--9422, 2020.

\bibitem[Regmi and Shah(2019)]{regmi}
Krishna Regmi and Mubarak Shah.
\newblock Bridging the domain gap for ground-to-aerial image matching.
\newblock In \emph{Proceedings of the IEEE/CVF International Conference on
  Computer Vision}, pages 470--479, 2019.

\bibitem[Sabour et~al.(2017)Sabour, Frosst, and Hinton]{capsnet}
Sara Sabour, Nicholas Frosst, and Geoffrey~E Hinton.
\newblock Dynamic routing between capsules.
\newblock In \emph{Proceedings of the 31st International Conference on Neural
  Information Processing Systems}, pages 3859--3869, 2017.

\bibitem[Shi et~al.(2019)Shi, Liu, Yu, and Li]{safa}
Yujiao Shi, Liu Liu, Xin Yu, and Hongdong Li.
\newblock Spatial-aware feature aggregation for image based cross-view
  geo-localization.
\newblock \emph{Advances in Neural Information Processing Systems},
  32:\penalty0 10090--10100, 2019.

\bibitem[Shi et~al.(2020{\natexlab{a}})Shi, Yu, Campbell, and
  Li]{where_looking}
Yujiao Shi, Xin Yu, Dylan Campbell, and Hongdong Li.
\newblock Where am i looking at? joint location and orientation estimation by
  cross-view matching.
\newblock In \emph{Proceedings of the IEEE/CVF Conference on Computer Vision
  and Pattern Recognition}, pages 4064--4072, 2020{\natexlab{a}}.

\bibitem[Shi et~al.(2020{\natexlab{b}})Shi, Yu, Liu, Zhang, and Li]{optimal}
Yujiao Shi, Xin Yu, Liu Liu, Tong Zhang, and Hongdong Li.
\newblock Optimal feature transport for cross-view image geo-localization.
\newblock In \emph{Proceedings of the AAAI Conference on Artificial
  Intelligence}, volume~34, pages 11990--11997, 2020{\natexlab{b}}.

\bibitem[Simonyan and Zisserman(2014)]{vgg}
Karen Simonyan and Andrew Zisserman.
\newblock Very deep convolutional networks for large-scale image recognition.
\newblock \emph{arXiv preprint arXiv:1409.1556}, 2014.

\bibitem[Sun et~al.(2019)Sun, Chen, Zhu, and Jiang]{geocaps}
Bin Sun, Chen Chen, Yingying Zhu, and Jianmin Jiang.
\newblock Geocapsnet: Ground to aerial view image geo-localization using
  capsule network.
\newblock In \emph{2019 IEEE International Conference on Multimedia and Expo
  (ICME)}, pages 742--747. IEEE, 2019.

\bibitem[Vaswani et~al.(2017)Vaswani, Shazeer, Parmar, Uszkoreit, Jones, Gomez,
  Kaiser, and Polosukhin]{transformer}
Ashish Vaswani, Noam Shazeer, Niki Parmar, Jakob Uszkoreit, Llion Jones,
  Aidan~N Gomez, Lukasz Kaiser, and Illia Polosukhin.
\newblock Attention is all you need.
\newblock \emph{arXiv preprint arXiv:1706.03762}, 2017.

\bibitem[Vo and Hays(2016)]{vo}
Nam~N Vo and James Hays.
\newblock Localizing and orienting street views using overhead imagery.
\newblock In \emph{European conference on computer vision}, pages 494--509.
  Springer, 2016.

\bibitem[Workman et~al.(2015)Workman, Souvenir, and Jacobs]{workman}
Scott Workman, Richard Souvenir, and Nathan Jacobs.
\newblock Wide-area image geolocalization with aerial reference imagery.
\newblock In \emph{Proceedings of the IEEE International Conference on Computer
  Vision}, pages 3961--3969, 2015.

\bibitem[Zhai et~al.(2017)Zhai, Bessinger, Workman, and Jacobs]{cvusa}
Menghua Zhai, Zachary Bessinger, Scott Workman, and Nathan Jacobs.
\newblock Predicting ground-level scene layout from aerial imagery.
\newblock In \emph{Proceedings of the IEEE Conference on Computer Vision and
  Pattern Recognition}, pages 867--875, 2017.

\bibitem[Zheng et~al.(2020)Zheng, Wei, and Yang]{university}
Zhedong Zheng, Yunchao Wei, and Yi~Yang.
\newblock University-1652: A multi-view multi-source benchmark for drone-based
  geo-localization.
\newblock In \emph{Proceedings of the 28th ACM international conference on
  Multimedia}, pages 1395--1403, 2020.

\end{thebibliography}


\end{document}